\documentclass{article}
\usepackage{mathptm}
\usepackage{colacl}
\usepackage{psfig}

\title{\vspace*{-10ex}{\small In {\it Proceedings of the 35th Annual Meeting of the Association for Computational Linguistics,} Madrid, 1997, pp.
507--509.} \\[7ex]
Choosing the Word Most Typical in Context \\ Using a Lexical Co-occurrence Network}

\author{Philip Edmonds\\
Department of Computer Science, University of Toronto \\
Toronto, Canada, M5S 3G4 \\
{\tt pedmonds@cs.toronto.edu}}

\newenvironment{rfigure*}
    {\begin{figure*}
       \hrule\vspace*{1ex}
   }{  \vspace*{1ex}\hrule
     \end{figure*}}

\newenvironment{rfigure}
    {\begin{figure}
       \hrule\vspace*{1ex}
   }{  \vspace*{1ex}\hrule
     \end{figure}}

\newenvironment{rtable*}
    {\begin{table*}
       \hrule\vspace*{1ex}
   }{  \vspace*{1ex}\hrule
     \end{table*}}

\newenvironment{rtable}
    {\begin{table}
       \hrule\vspace*{1ex}
   }{  \vspace*{1ex}\hrule
     \end{table}}

\newcommand{\listlengths}[1]{
    \settowidth{\labelwidth}{#1}
    \setlength{\labelsep}{1.0em}
    \setlength{\leftmargin}{\labelwidth}
    \addtolength{\leftmargin}{\labelsep}}

\newcounter{sentence}
\renewcommand{\thesentence}{(\arabic{sentence})}
\newenvironment{sentence}
   {\begin{list}{\thesentence}{\listlengths{(MM)}}
   }{\end{list}}
\newcommand{\sent}{\refstepcounter{sentence}\item\raggedright}

\begin{document}

\maketitle 
\vspace{-0.5in}
\begin{abstract}
  This paper presents a partial solution to a component of the problem
  of lexical choice: choosing the synonym most typical, or expected,
  in context.  We apply a new statistical approach to representing the
  context of a word through lexical co-occurrence networks.  The
  implementation was trained and evaluated on a large corpus, and
  results show that the inclusion of second-order co-occurrence
  relations improves the performance of our implemented lexical choice
  program.
\end{abstract}

\section{Introduction}

Recent work views lexical choice as the process of mapping from a set
of concepts (in some representation of knowledge) to a word or phrase
\cite{elhadad-thesis,stede-thesis}.  When the same concept admits more
than one lexicalization, it is often difficult to choose which of
these `synonyms' is the most appropriate for achieving the desired
pragmatic goals; but this is necessary for high-quality machine
translation and natural language generation.

Knowledge-based approaches to representing the potentially subtle
differences between synonyms have suffered from a serious lexical
acquisition bottleneck \cite{dimarco93,hirst95}.  Statistical
approaches, which have sought to explicitly represent differences
between pairs of synonyms with respect to their occurrence with other
specific words \cite{church94}, are inefficient in time and
space.

This paper presents a new statistical approach to modeling context
that provides a preliminary solution to an important sub-problem, that
of determining the near-synonym that is {\em most typical}, or
expected, if any, in a given context.  Although weaker than full
lexical choice, because it doesn't choose the `best' word, we believe
that it is a necessary first step, because it would allow one to
determine the effects of choosing a non-typical word in place of the
typical word.  The approach relies on a generalization of lexical
co-occurrence that allows for an implicit representation of the
differences between two (or more) words with respect to {\em any}
actual context.

For example, our implemented lexical choice program selects {\em
  mistake} as most typical for the `gap' in sentence~\ref{mistake72},
and {\em error} in \ref{error115}.
\begin{sentence}
  \sent\label{mistake72} However, such a move also would run the
  risk of cutting deeply into U.S. economic growth, which is why some
  economists think it would be a big \{error $|$ mistake $|$ oversight\}.
  
  \sent\label{error115} The \{error $|$ mistake $|$ oversight\}\ was
  magnified when the Army failed to charge the standard percentage
  rate for packing and handling.
\end{sentence}

\section{Generalizing Lexical Co-occurrence}

\subsection{Evidence-based Models of Context}

Evidence-based models represent context as a set of features, say
words, that are observed to co-occur with, and thereby predict, a word
\cite{yarowsky92,golding-schabes96,karow-edelman96,ng-lee96}.  But, if
we use just the context surrounding a word, we might not be able to
build up a representation satisfactory to uncover the subtle
differences between synonyms, because of the massive volume of text
that would be required.

Now, observe that even though a word might not co-occur significantly
with another given word, it might nevertheless {\em predict} the use
of that word if the two words are mutually related to a third word.
That is, we can treat lexical co-occurrence as though it were
moderately transitive.  For example, in \ref{task113}, {\em learn}
provides evidence for {\em task} because it co-occurs (in other
contexts) with {\em difficult}, which in turn co-occurs with {\em
  task} (in other contexts), even though {\em learn} is not seen to
co-occur significantly with {\em task}.
\begin{sentence}
  \sent\label{task113} {The team's most urgent {\em task} was to {\em
      learn} whether Chernobyl would suggest any safety flaws at
    KWU-designed plants.}
\end{sentence}
So, by augmenting the contextual representation of a word with such
second-order (and higher) co-occurrence relations, we stand to have
greater predictive power, assuming that we assign less weight to them
in accordance with their lower information content.  And as our
results will show, this generalization of co-occurrence is necessary.

We can represent these relations in a {\em lexical co-occurrence
  network}, as in figure~\ref {task-net}, that connects lexical items
by just their first-order co-occurrence relations.  Second-order and
higher relations are then implied by transitivity.

\begin{rfigure}
\psfig{file=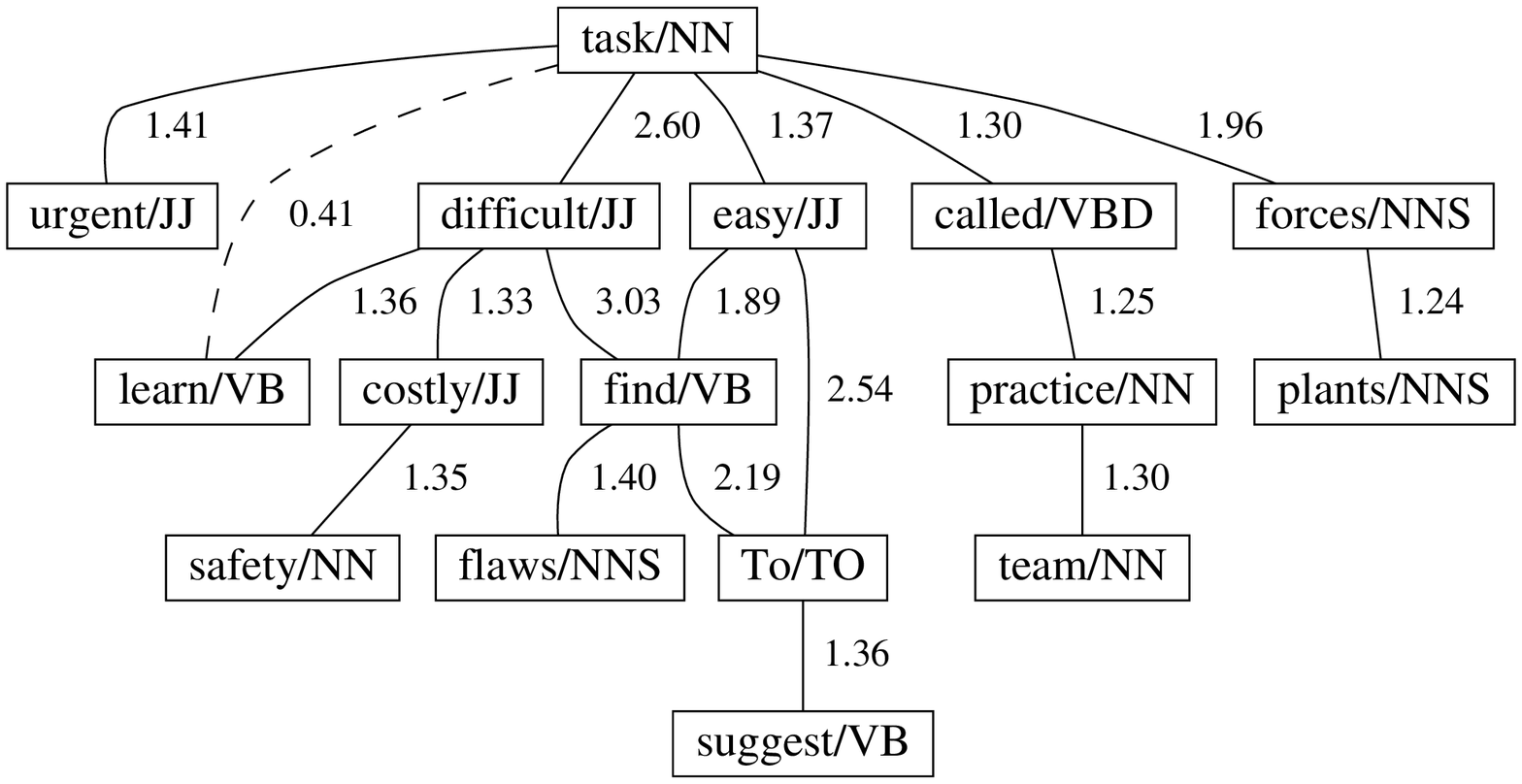,width=3.1in}
\caption{A fragment of the lexical co-occurrence network for {\em
    task}.  The dashed line is a second-order relation implied by the
  network.}\label{task-net}
\end{rfigure}

\subsection{Building Co-occurrence Networks}

We build a lexical co-occurrence network as follows: Given a root
word, connect it to all the words that significantly co-occur with it
in the training corpus;\footnote{Our training corpus was the
  part-of-speech-tagged 1989 {\it Wall Street Journal}, which consists
  of $N = 2,709,659$ tokens.  No lemmatization or sense disambiguation
  was done.  Stop words were numbers, symbols, proper nouns, and any
  token with a raw frequency greater than $F=800$.}  then, recursively
connect these words to their significant co-occurring words up to some
specified depth.

We use the intersection of two well-known measures of significance,
mutual information scores and $t$-scores \cite{church94}, to determine
if a (first-order) co-occurrence relation should be included in the
network; however, we use just the $t$-scores in computing {\em
  significance scores} for all the relations.  Given two words, $w_0$
and $w_d$, in a co-occurrence relation of order $d$, and a shortest
path $P(w_0,w_d) = (w_0,\ldots,w_d)$ between them, the significance
score is
\[
\mbox{sig}(w_0,w_d) = \frac{1}{d^3}\sum_{w_i \in P(w_1,w_d)}{\frac{t(w_{i-1},w_i)}{i}}
\]
This formula ensures that significance is inversely proportional to the
order of the relation.  For example, in the network of
figure~\ref{task-net}, sig$(\mbox{\em task}, \mbox{\em learn}) = [
t(\mbox{\em task}, \mbox{\em difficult}) + \frac{1}{2}t(\mbox{\em
  difficult}, \mbox{\em learn} ) ]/8 = 0.41$.

A single network can be quite large.  For instance, the complete
network for {\em task} (see figure~\ref{task-net}) up to the
third-order has 8998 nodes and 37,548 edges.

\subsection{Choosing the Most Typical Word}

The amount of evidence that a given sentence provides for choosing a
candidate word is the sum of the significance scores of each
co-occurrence of the candidate with a word in the sentence.
So, given a gap in a sentence $S$, we find the candidate $c$ for
the gap that maximizes
\[ 
 M(c,S) = \sum_{w \in S}{\mbox{sig}(c,w)}
\]
For example, given $S$ as sentence~\ref{task113}, above, and the network of
figure~\ref{task-net}, $M(\mbox{\em task},S) = 4.40$.  However, {\em
  job} (using its own network) matches best with a score of $5.52$;
{\em duty} places third with a score of $2.21$.

\section{Results and Evaluation}

To evaluate the lexical choice program, we selected several sets of
near-synonyms, shown in table~\ref{syn-sets}, that have low polysemy
in the corpus, and that occur with similar frequencies.  This is to
reduce the confounding effects of lexical ambiguity.

\begin{rtable}
\begin{center}
\small

\begin{tabular}{lll}
Set & POS & Synonyms (with training corpus frequency)\\
\hline
1 & {\sc jj} & difficult (352),  hard (348), tough (230) \\
2 & {\sc nn} & error (64), mistake (61), oversight (37) \\ 
3 & {\sc nn} & job (418), task (123), duty (48) \\
4 & {\sc nn} & responsibility (142), commitment (122), \\
  & &    obligation (96), burden (81) \\
5 & {\sc nn} & material (177), stuff (79), substance (45) \\
6 & {\sc vb} & give (624), provide (501), offer (302) \\ 
7 & {\sc vb} & settle (126), resolve (79) \\
\hline
\end{tabular}

\end{center}
\caption{The sets of synonyms for our experiment.}\label{syn-sets}
\end{rtable}

For each set, we collected all sentences from the yet-unseen 1987 {\it
  Wall Street Journal} (part-of-speech-tagged) that contained any of
the members of the set, ignoring word sense.  We replaced each
occurrence by a `gap' that the program then had to fill.  We compared
the `correctness' of the choices made by our program to the baseline
of always choosing the most frequent synonym according to the training
corpus.

But what are the `correct' responses?  Ideally, they should be chosen
by a credible human informant.  But regrettably, we are not in a position
to undertake a study of how humans judge typical usage, so we will
turn instead to a less ideal source: the authors of the {\it
  Wall Street Journal}.  The problem is, of course, that authors
aren't always typical.  A particular word might occur in a `pattern'
in which another synonym was seen more often, making it the typical
choice.  Thus, we cannot expect perfect accuracy in this evaluation.

Table~\ref{table-results} shows the results for all seven sets of
synonyms under different versions of the program. We varied two
parameters: (1) the window size used during the construction of the
network: either narrow ($\pm$4 words), medium ($\pm$ 10 words), or
wide ($\pm$ 50 words); (2) the maximum order of co-occurrence relation
allowed: 1, 2, or 3.  

\begin{rtable*}
\begin{center}

\begin{tabular}{rrrrrrrr}
\hline
Set & 1 & 2 & 3   & 4 & 5 & 6 & 7 \\
Size & 6665 & 1030 & 5402   & 3138 & 1828 & 10204 & 1568 \\[0.5ex]
Baseline & 40.1\% & 33.5\% & 74.2\%   & 36.6\% & 62.8\% & 45.7\% & 62.2\% \\[0.5ex]

       1 & 31.3\% & 18.7\% & 34.5\%   & 27.7\% & 28.8\%  & 33.2\% & 41.3\% \\
Narrow 2 & 47.2\% & 44.5\% & 66.2\%   & 43.9\% & 61.9\%$^a$ & 48.1\% & 62.8\%$^a$ \\
       3 & {\bf 47.9\%} & {\bf 48.9\%} & {\bf 68.9\%}   & 44.3\% & {\bf 64.6\%$^a$} & {\bf 48.6\%} & {\bf 65.9\%} \\[0.5ex]

       1 & 24.0\% & 25.0\% & 26.4\%   & 29.3\% & 28.8\% & 20.6\% & 44.2\% \\
Medium 2 & 42.5\% & 47.1\% & 55.3\%   & {\bf 45.3\%} & 61.5\%$^a$ & 44.3\% & 63.6\%$^a$ \\
       3 & 42.5\% & 47.0\% & 53.6\%   & --- & --- & --- & --- \\[0.5ex]

Wide 1 & 9.2\% & 20.6\% & 17.5\%      & 20.7\% & 21.2\% & 4.1\% & 26.5\% \\ 
     2 & 39.9\%$^a$ & 46.2\% & 47.1\%    & 43.2\% & 52.7\% & 37.7\% & 58.6\% \\
\hline 
\multicolumn{8}{l}{\footnotesize $^a$Difference from baseline not significant.}
\end{tabular}

\end{center}

\caption{Accuracy of several different versions of the lexical choice
  program.  The best score for each set is in boldface.  Size refers
  to the size of the sample collection.  All differences from baseline are
  significant at the $5\%$ level according to Pearson's $\chi^2$ test,
  unless indicated.}\label{table-results}
\end{rtable*}

The results show that at least second-order co-occurrences are
necessary to achieve better than baseline accuracy in this task;
regular co-occurrence relations are insufficient.  This justifies our
assumption that we need more than the surrounding context to build
adequate contextual representations.

Also, the narrow window gives consistently higher accuracy than the
other sizes.  This can be explained, perhaps, by the fact that
differences between near-synonyms often involve differences in
short-distance collocations with neighboring words, e.g., {\em face
  the task}.

There are two reasons why the approach doesn't do as well as an
automatic approach ought to.  First, as mentioned above, our method of
evaluation is not ideal; it may make our results just {\em seem} poor.
Perhaps our results {\em actually} show the level of `typical usage'
in the newspaper.

Second, lexical ambiguity is a major problem, affecting both
evaluation and the construction of the co-occurrence network.  For
example, in sentence \ref{task113}, above, it turns out that the
program uses {\em safety} as evidence for choosing {\em job} (because
{\em job safety} is a frequent collocation), but this is the wrong
sense of {\em job}.  Syntactic and collocational red herrings can add
noise too.

\section{Conclusion}

We introduced the problem of choosing the most typical synonym in
context, and gave a solution that relies on a generalization of
lexical co-occurrence.  The results show that a narrow window of
training context ($\pm4$ words) works best for this task, and that at
least second-order co-occurrence relations are necessary.  We are
planning to extend the model to account for more structure in the
narrow window of context.

\section*{Acknowledgements}

For comments and advice, I thank Graeme Hirst, Eduard Hovy, and
Stephen Green.  This work is financially supported by the Natural
Sciences and Engineering Council of Canada.


\small
\bibliographystyle{acl}

\end{document}